\documentclass{article}

    \usepackage[final]{neurips_2023}

\usepackage[utf8]{inputenc} 
\usepackage[T1]{fontenc}    
\usepackage{hyperref}       
\usepackage{url}            
\usepackage{booktabs}       
\usepackage{amsfonts}       
\usepackage{nicefrac}       
\usepackage{microtype}      
\usepackage{xcolor}         

\usepackage{svg}
\usepackage{indentfirst}
\usepackage{bm}
\usepackage{amssymb}
\usepackage{amsmath}
\usepackage{natbib}
\usepackage{caption}
\title{Emergent Explainability: Adding a causal chain to neural network inference}

\author{%
  Adam Perrett\thanks{Corresponding author} \\
  The University of Manchester\\
  Manchester, UK \\
  \texttt{adam.perrett@manchester.ac.uk} \\
}

\begin{document}

\maketitle

\begin{abstract}
This position paper presents a theoretical framework for enhancing explainable artificial intelligence (xAI) through emergent communication (EmCom), focusing on creating a causal understanding of AI model outputs. We explore the novel integration of EmCom into AI systems, offering a paradigm shift from conventional associative relationships between inputs and outputs to a more nuanced, causal interpretation. The framework aims to revolutionize how AI processes are understood, making them more transparent and interpretable. While the initial application of this model is demonstrated on synthetic data, the implications of this research extend beyond these simple applications. This general approach has the potential to redefine interactions with AI across multiple domains, fostering trust and informed decision-making in healthcare and in various sectors where AI's decision-making processes are critical. The paper discusses the theoretical underpinnings of this approach, its potential broad applications, and its alignment with the growing need for responsible and transparent AI systems in an increasingly digital world.

\end{abstract}
\section{Introduction and background}
\subsection{Opportunities and challenges of explainability}
Explainable artificial intelligence (xAI) remains one of the foremost challenges in the field, crucial to equitable and fair application, particularly in sensitive domains like healthcare. 
For examples, understanding the impact of race in AI models is a common problem\cite{Huang2022racialbias}. Mitigating potential bias that exists within data and its effect on treatment begins with understanding the AI model. 
The core issue lies in the inherent opacity of advanced AI models, especially deep learning algorithms, which operate as "black boxes". 
These complex models can make highly accurate predictions or decisions, but understanding the causal chain leading to their output is not straightforward. This lack of transparency raises concerns about trustworthiness, accountability, and ethics, especially when AI is employed in critical areas such as medical diagnostics, treatment planning, or patient care management. 
This work addresses concerns of xAI with the goal of enabling AI to be a vital and interpretable tool for clinicians and patients alike, fostering trust and enabling informed decision-making. The novel framework also has many potential applications beyond healthcare, with the potential to reshape the way we interact with AI.


In the healthcare sector, xAI methods face significant challenges. To address concerns of safety, equity and trustworthiness, an understanding of causality is crucial. Simpler models like decision trees trade complexity for clarity, essential where decision rationales are crucial. However, complex models using tools like LIME~\cite{Ribeiro2016LIME} and SHAP~\cite{Lundberg2017SHAP} often yield limited, sometimes misleading insights. Visualization techniques in deep learning, such as attention maps, provide helpful interpretations but lack precise correlations with neural network processes~\cite{Erhan2009visGA, Simonyan2013visGAvsdeconv}. Feature importance measures, while identifying key decision factors, can suffer from bias and lack causal clarity~\cite{Zhou2016saliency, Selvaraju2017gradcam}. These methods, enhancing reliability and aiding compliance, struggle with a fundamental trade-off between accuracy and explainability. This leads to technical complexities and the risk of oversimplified interpretations, underscoring the need for more refined and transparent explainability techniques in high-stakes healthcare AI applications. 

This work focuses on enhancing xAI in healthcare, a key area in need of growth and equitable application. It aims to develop a novel and practical framework for AI explainability, particularly in healthcare diagnostics and patient care. 
The proposed research will imbue the AI model with a causal communication channel that can be used for validation and transparency of output.
The expected outcomes include improved AI clarity and trust between clinicians and patients, with the potential to inform policy in AI ethics. This effort is geared towards contributing meaningful research and reshaping the way we use AI as a tool.

\begin{figure}
    \centering
    \includegraphics[width=1.\textwidth]{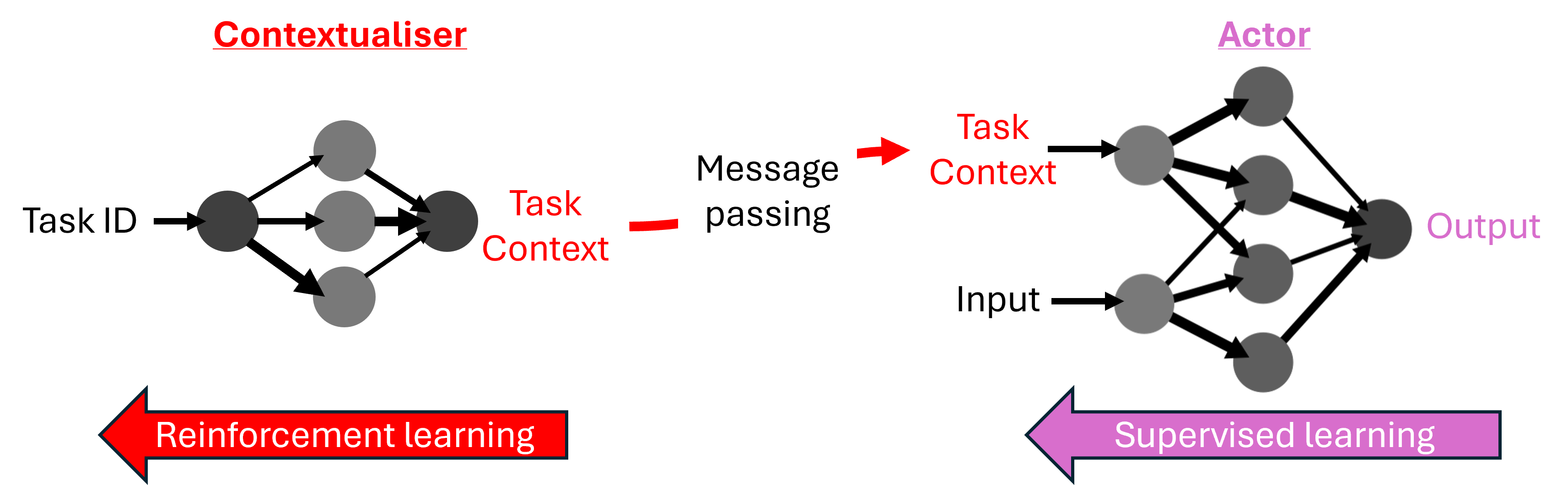} 
    \captionsetup{width=0.48\textwidth,font={small,stretch=0.80},justification=raggedright,singlelinecheck=false,margin={0pt,0pt}}
    \caption{The contextualiser network receives the task ID (e.g. task = is there a dog) and passes a message to the actor network. The actor network processes the message and an input to produce an output (e.g. is there a dog + this image = yes/no). The actor is trained with supervised learning and the contextualiser is trained with reinforcement learning, as errors cannot propagate between networks.}
    \label{fig:communication}
\end{figure}
\vspace{0pt}

\subsection{Emergent communication}
Emergent communication (EmCom) in AI~\cite{Brandizzi2023communicationreview}, a relatively nascent field, presents unique challenges and opportunities in the context of explainability. The principle behind EmCom is to have two or more agents exchanging messages in order to solve a task, see Fig.\ref{fig:communication}. Altering the task set-up can alter the compositionality and generalisation of the generated language~\cite{Mu2021EmerGeneralisation, Lowe2020SupervisedComm}. Current research focuses on investigating the generated language and comparing it to human language~\cite{Li2019TeachingEmergentComm, Lazaridou2020MAcommunication, Lowe2019EmergentComm}. The research community sees the opportunity to apply this further and scale it up~\cite{chaabouni2022emergent}, highlighting the growth possible in this fruitful domain. A key shift this work aims to exploit is moving from message passing being input dependant to task dependant. This raises the level of abstraction of the message from a label to an instruction.

\subsection{Emergent explainability}
The aim of this research is to utilise the mechanisms of EmCom to make neural network explainability causal. Current xAI methods reveal associative relationships between inputs and outputs but are often limited to analysing single input presentations and say little about causality. The proposed work hinges on the following principles:
\begin{itemize}
    \item The messages used in EmCom can be made human interpretable (e.g. text or image)
    \item The message and the output share a causal relationship, as the task can only be solved with the message
    \item An appropriately generalised language facilitates: transfer learning, data privatisation, distributed learning
\end{itemize}

\section{Methodology}
Experimental work focuses on synthetic data. Work has also been performed using the MNIST dataset, although this is still currently under investigation. Preliminary results will be discussed, with detailed analysis to follow. Integration with human interpretable message passing is currently theoretical. 
\subsection{Experimental design}
All possible truth tables for three inputs ($n=3$) act as the family of tasks that are trained on. This gives $2^{2^3} = 256$ possible truth tables and $2^3 * 2^{2^3} = 1024$ training examples. Each truth table is allocated a random One Hot Encoding (OHE) to act as the task ID, this will be passed to the contextualiser to generate a truth table specific message. A random OHE is used to unsure there is as little possible information present in the task ID and all communicated information must be learnt during training. The actor will then receive the message and a set of 3 inputs. 
\subsection{Agent setup}
Each agent has a separate contextualiser network and actor network. Both have two hidden layers of 128 ReLU activation neurons, only difference between them being their inputs, outputs, and initialisation. The conextualiser has as many inputs as there are tasks, as it receives a OHE for the task ID. There is a context layer of 32 ReLU neurons places after the final hidden layer. The activation of these neurons forms the message that is passed. There is then a single dense layer which samples from this to predict the expected reward of the actor. The error between this and the actual actor reward forms the reinforcement learning signal. The actor network has $3 + 32$ inputs (input size + context length) and 2 outputs (binary 0 and 1 for the truth table entry). It is trained using supervised cross-entropy.
\subsection{Communication protocol}
\begin{figure*}
    \centering
    \includegraphics[width=1.\textwidth]{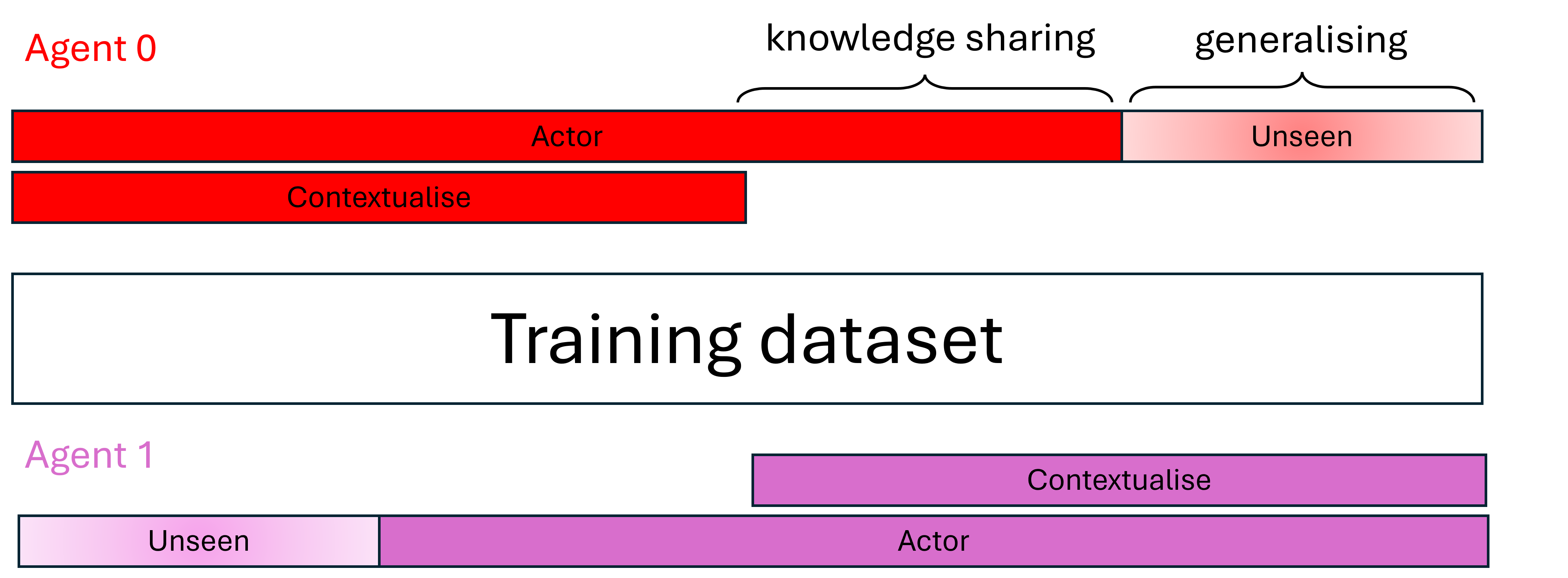} 
    \captionsetup{width=0.48\textwidth,font={small,stretch=0.80},justification=raggedright,singlelinecheck=false,margin={0pt,0pt}}
    \caption{An agent can act in three different capacities during training. It can contextualise, meaning it takes the task ID and produces a message. It can behave as the actor, meaning it uses the message and the input to produce the target output during these training instances. It can be an actor for examples it contextualises and ones other agents contextualise. There is then a subset of data that is never seen by the agent, this is used to evaluate the generalisation of the generated language.}
    \label{fig:teaching}
\end{figure*}

Fig.~\ref{fig:teaching} shows how the training data is distributed. Agents are allocated a unique portion of tasks which it can act as a contextualiser for, $r_c$. It can act as an actor for all these and an additional portion of the remaining tasks, $r_a$. The remaining tasks will be unseen by the agent throughout training. 
There is a truth table associated with each training example. For each example an agent which can act as a contextualiser is randomly selected. It is given as input the OHE associated with the task ID and produces the associated context. An agent which can be an actor for this training example is then selected and given the associated message and input. 
\subsection{Training}
A batch size of 512 is used with an Adam optimiser using learning rate 0.001. Cross-entropy loss between the target and actual output is used to train the actor. Mean-squared error between the predicted and actual actor accuracy is used to train the contextualiser using reinforcement learning. When the contextualiser and actor are both the same agent, gradients are allowed to flow back through the actor and into the communicated message.
\vspace{0pt}
\subsection{Evaluation}
The key evaluation metric is the performance of an agent on unseen truth tables. The only way to achieve higher than random chance accuracy is for the communicated messages to contain generalised information about the structure of the family of tasks. This way, even though an agent has never performed the task, it can understand the message and perform the task adequately. 
\subsection{Work in Progress - Human interpretability}
This methodology is currently under development. The principle is to alter the message passing between agents to be human interpretable. This can take the form of an image related to the task, such as the features to look for e.g. the task is looking for a dog so the communicated message contains a nose and floppy ears. This can be injected into the convolutional kernels or concatenated downstream. It is also possible that the message passing can be human language. This would require the message to contain all relevant information to complete the task.

\section{Experiments and Results}
\subsection{Examining the effect of actor overlap}
\begin{figure*}
    \centering
    \includegraphics[width=1.\textwidth]{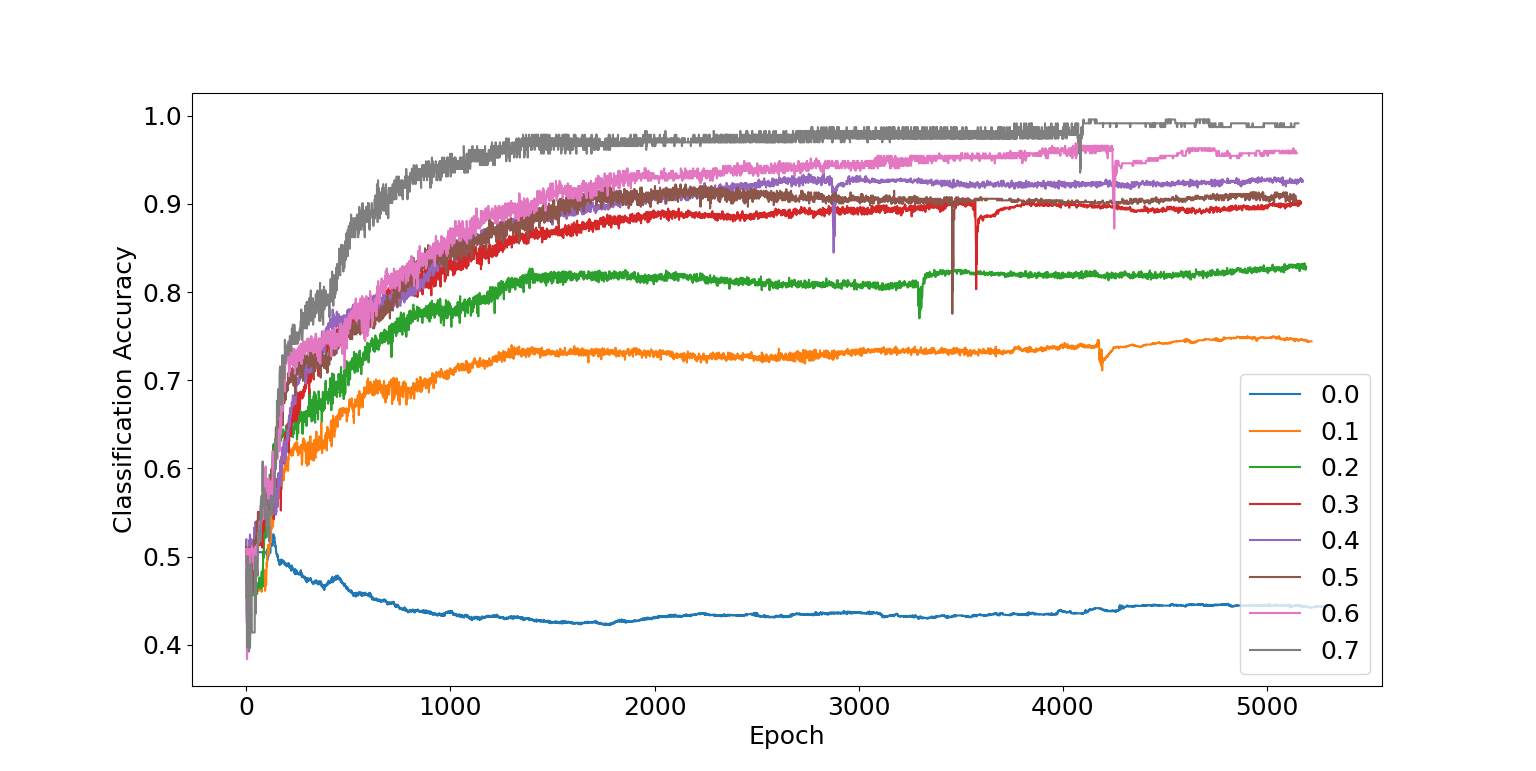} 
    \captionsetup{width=0.48\textwidth,font={small,stretch=0.80},justification=raggedright,singlelinecheck=false,margin={0pt,0pt}}
    \caption{Classification accuracy on unseen data across all agents for n=3 truth table size. The number of example that are contextulised by an agent is fixed and the parameter $r_a$, determining the number of examples that an agent is an actor for but not a contextualiser, is investigated. It can be seen that as the number of 'taught' examples increased the generalisation of the communication also increases, tending towards 100\% testing accuracy.}
    \label{fig:overlap_n3}
\end{figure*}

The variable $r_a$, controlling the ratio of examples in which an agent is not a contextualiser but can be an actor is explored in Fig,~\ref{fig:overlap_n3}. As the amount of 'information sharing' increases so does the classification accuracy on unseen data. This displays that interaction between agents is key to generating a generalised communication. This is most evident when there is no sharing and the performance does not increase above random chance. 
\vspace{0pt}

\section{Discussion and Limitations}
This work forms a position paper displaying the potential for increasing abstract message passing between agents compared to the current literature. It has been displayed for communicating information about a family of tasks and has been shown to generalise outside of tasks an agent is trained on.

There is still much work to do in imbuing this communication channel with human interpretable messages but this is a stepping stone in that direction and a signpost for future endeavours.

With the novel transition in this work of EmCom from communicating a label to communicating task information, the language is a higher level of abstraction and more focused on function. It has already been proven that languages can emerge which generalise beyond training data~\cite{Mu2021EmerGeneralisation, ?}. This naturally leads to networks which can generalise across tasks, given the appropriate context. This can allow training data to remain proprietary whilst still sharing learnt information with another network. It can enable distributed learning, as many agents can go off and learn different tasks and return with their understanding. Obvious extensions to transfer learning exist due to the task independent nature of the language. There is also reason to believe that embedding such causal grounding in LLMs will help alleviate problems such as hallucinations and mistakes with trivial request like simple maths. This is because the framework will remove the sole dependence on the statistical regularities of text and instead reposition the text as a medium with which the task can be understood. When this condition is met the initial contextualiser in the EmCom framework can be replaced with a human, and the actor can then be harnessed as a generalised task solver.






\medskip

{
\small

\bibliography{myrefs}


\end{document}